\documentclass[11pt]{article}
\usepackage{booktabs}
\usepackage{caption}
\usepackage{subcaption}
\usepackage[utf8]{inputenc}
\usepackage[T1]{fontenc}
\usepackage{url}
\usepackage{booktabs}
\usepackage{amsfonts}
\usepackage{nicefrac}
\usepackage{amsmath}
\usepackage{booktabs}
\usepackage{xspace}
\usepackage{graphicx}
\usepackage{bm}
\usepackage{hyperref}
\usepackage{natbib}
\usepackage{rotating}
\usepackage[margin=1in]{geometry}
\usepackage{setspace}
\usepackage{tabularx}
\usepackage{ragged2e}
\usepackage{authblk}\usepackage{abstract}
\usepackage{lineno}

\title{\textbf{Who Delegates to AI? Evidence from Agent Configurations in Github}}

\author[1]{Hyeongjae Lee}
\author[2]{Jihyang Cheon}
\author[1,2]{Lanu Kim$^{\dagger}$}
\affil[1]{Graduate School of Digital Humanities and Social Sciences,
  Korea Advanced Institute of Science and Technology (KAIST),
  Daejeon, Republic of Korea}
\affil[2]{Graduate School of Data Science,
  Korea Advanced Institute of Science and Technology (KAIST),
  Daejeon, Republic of Korea}

\begin{document}
\maketitle

\begin{abstract}

A growing body of literature measures the extent to which occupations are exposed to AI, yet existing measures capture where AI could perform tasks rather than whether workers have actually adopted it. We introduce a distinct tier of exposure, delegated exposure, which records whether a worker has committed a task to AI by embedding it into a structured workflow. We operationalize this concept through the Agentic Adoption Index (AAI), measuring how closely an occupation's tasks align with the agentic routines that practitioners have built and shared. Using semantic embeddings of roughly 888,000 agent skill specifications from public GitHub repositories, we compute their similarity to nearly 18,000 O*NET task statements and aggregate these scores to the occupational level. We present three main findings. First, the occupations where task delegation concentrates differ sharply from those identified as most vulnerable by pre-AI automation frameworks. Second, the AAI aligns more closely with measures of technical capability than with measures of current conversational LLM use. Third, for occupations requiring a bachelor's degree or less, the AAI increases alongside average wage levels; however, this relationship reverses for occupations requiring a master's degree or higher, where adoption declines among higher earners. These patterns replicate on an independently collected corpus of agent skills from the Manus Skills Marketplace. This lower adoption among highly educated, high-earning workers may reflect tasks that inherently resist advance specification or professional discretion over the pacing of workflow codification. Distinguishing these mechanisms will require longitudinal measurement.

\end{abstract}

\section{\label{sec:intro}Introduction}

While artificial intelligence now sits at the center of technological change, accurately measuring its labor market impacts remains an ongoing challenge. Several factors compound this difficulty: the technology is evolving rapidly, its reach is exceptionally broad—extending to unstructured tasks across diverse occupations, and we still lack a clear picture of who is adopting these tools and for what purposes \citep{bick2026rapid, frank2026aiexposedjobsdeterioratedchatgpt}. These characteristics make AI's labor-market impacts harder to predict than those of earlier automation \citep{frank2019labor}, motivating a growing body of research that measures how exposed the tasks within occupations are to AI, and more recently, how far that exposure has translated into use.

We group the extensive literature on occupational exposure to AI or automation technology into three lines of research, ordered by how closely each comes to observing the realized adoption of AI. The first assesses which tasks AI is capable of performing, then identifies the occupations in which those capabilities would be useful \citep{felten_method_2018, felten_effect_2019}; what it measures is a technological ceiling. The second narrows that ceiling to what firms have chosen to build and sell: the commercially available applications through which large language models actually reach users \citep{felten_how_2023, eloundou2024gpts}. The third narrows it again, to what workers choose to use, examining which work-related tasks actually appear in people's interactions with conversational LLMs \citep{handa2025economic, anthropic2026aeiv5, tomlinson_working_2025}. Despite the variety of methods, measures, and data they employ, these studies converge: whereas earlier automation displaced routine work, AI reaches cognitively demanding work \citep{frank2019labor, eloundou2024gpts, handa2025economic}.

These three lines share a common limit: each measures how far AI could or does touch an occupation's work, not whether a worker has committed the work to it. We propose a fourth, which we call \textit{delegated exposure}. What technology — or commercialized technology — is capable of has grown distant from what people actually adopt in practice \citep{delriochanona2025aijobsreviewtheory}, and the gap matters especially now that the technology is moving toward agentic systems: a recent class of LLM-based systems that pursue specified goals and carry out recurring activities through structured workflows \citep{yang2025adoption, gupta2026agenticaioccupationaldisplacement, xing2026critiqueagentmodel}. Setting up such a system requires practitioners to define its goals, instructions, and workflow in advance, and because these configurations automate complex, multi-step work, each one records a deliberate choice to deploy AI for a specific task. That is a different kind of evidence from a conversational log: it shows not just that a worker used AI, but that they delegated a task to it. We therefore ask where delegated exposure is concentrating in the occupational structure, and how that distribution compares with what the three existing measures would predict.


To measure this, we observe adoption directly, measuring how much of each occupation's task content practitioners have already built agents for. We construct an Agentic Adoption Index (AAI) from GitSkills \citep{destefanis2026gitskillsdatasetagentskills}, an open dataset of agent \texttt{skill} files collected from public GitHub repositories, the only publicly accessible source at scale where practitioners share the \texttt{skill} files that configure their agents. Using Sentence-BERT (all-MiniLM-L6-v2), we convert roughly 888,000 agent \texttt{skill} specifications into embeddings, vector representations that capture their meaning. We then compute the semantic similarity between these embeddings and those of about 18,000 O*NET task statements, and aggregate the resulting similarities to the occupation level, weighting each task by its importance within the occupation. The AAI therefore captures how closely an occupation's tasks match the agentic routines that practitioners have already built and shared, and is designed for comparison across occupations rather than interpretation on an absolute scale. Contributors are practitioners who both build agents and make them public, so the measure reflects tasks whose routines generalize well enough to be worth sharing. Since the population of public \texttt{skill}s grows continuously, the same procedure can be repeated to capture shifts in what practitioners choose to automate, offering a way to track adoption as it unfolds rather than only in retrospect. Such tracking may serve as an early indicator of where AI is beginning to reshape work, to the extent that agent configurations appear before their impacts register in employment statistics. 

Our analysis of the AAI yields three findings. First, the occupations where agent adoption concentrates differ sharply from those that earlier automation research identified as most exposed. Second, the AAI broadly aligns more closely with measures of what AI is technically capable of than with observed patterns of conversational LLM use. Because building an agent requires deliberate configuration, practitioners invest only where the technology can reliably do the work, whereas conversational use is cheap enough to spread without regard to capability. Third, this capability-driven alignment breaks down at the top of the labor market: adoption falls off sharply among the highest-paid and most-educated occupations. Although technical availability accounts for some variance across income and education levels, it cannot account for the lower rate of adoption observed at the upper end of these distributions. Technical feasibility is therefore a necessary condition for agentic adoption, but not a sufficient one to explain where it occurs. 

This paper contributes a way to measure the adoption of agentic AI as it happens, rather than inferring it from what the technology can do — a distinction that matters because capability does not guarantee use. The measure has two practical advantages. Because it draws on a continuously growing repository, it can be recomputed as adoption evolves, unlike exposure measures anchored to periodic expert assessments. And because \texttt{skill} files are artifacts practitioners built for their own work, the measure captures revealed preference rather than judgments about what AI could plausibly do. Substantively, our findings speak to the sociology of professions: the occupations best positioned to automate their work are not the ones doing so, suggesting that adoption at the top of the labor market is governed by something other than technical feasibility.

\section{\label{sec:background}Background and Related Work}
\subsection{\label{sec:exposure}Four layers of AI exposure}

Occupations are not single, uniform jobs: each comprises a distinct bundle of tasks and draws on different skills and abilities. This matters because exposure varies considerably across the tasks within a single occupation, so a measure that treats jobs as indivisible units can obscure which parts of the work are actually at stake. Research on AI's impact at work has therefore converged on a common strategy, made possible by O*NET, a well-established dataset that specifies these elements in fine-grained form. Studies assess how far individual tasks \citep{eloundou2024gpts, anthropic2026aeiv5}, work activities \citep{brynjolfsson_what_2018, tomlinson_working_2025}, or skills \citep{felten_method_2018, felten_occupational_2021, chopra2025icebergindexmeasuringskillscentered} will be affected by AI, then aggregate those assessments to the occupation level. We follow this strategy in our analysis. 

\begin{figure}[t]
    \centering
    \includegraphics[width=\linewidth]{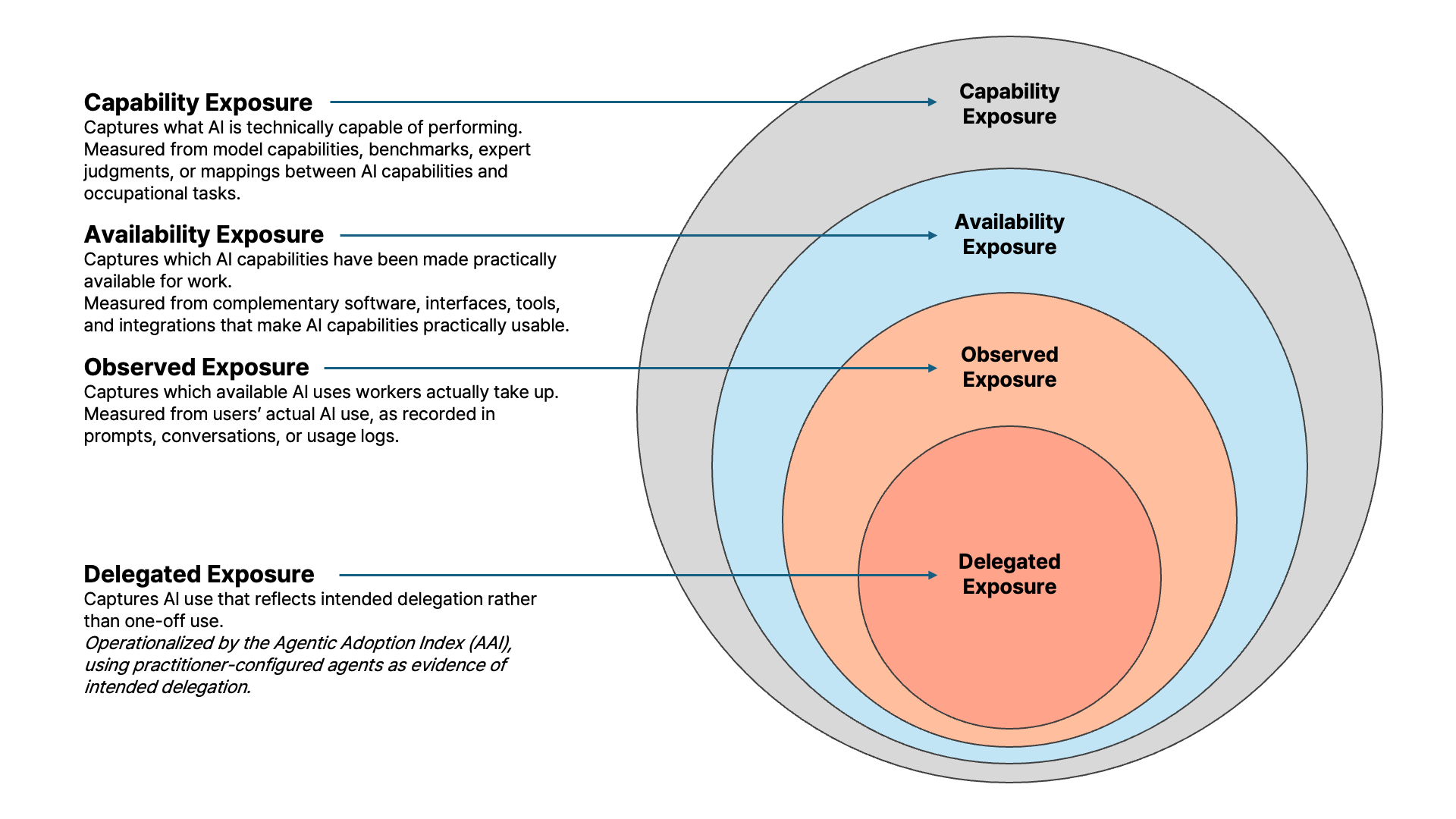}
    \caption{Four nested layers of AI exposure: \textit{capability} (what AI can do), \textit{availability} (what is made usable), \textit{observed} (what workers use), and \textit{delegated} (what workers commit to AI). Each layer is nested within the one before it.}
    \label{fig:exposure_layers}
\end{figure}

While sharing this approach, the literature diverges on which technology it measures. Mouchel et al. \citep{mouchel2026jobsaiexposuremeasured} and Yin and Ogut \citep{yin2026usesaiplatformselection} each distinguish capability and availability from use, but neither specifies availability from capability. We therefore organize this research into three layers of AI's reach: what AI could technically do, what commercial tools make usable, and what people actually do with it (Fig.~\ref{fig:exposure_layers}). The layers are approximately nested, since not every technical capability is built into an available tool, and not every available tool is ultimately used. We review each in turn.

The first layer measures what AI could technically do, in order to identify which occupations it would affect. Brynjolfsson, Mitchell, and Rock rated individual O*NET tasks and confirmed that tasks within the same occupation often differ sharply in their susceptibility to AI \citep{brynjolfsson_what_2018}. Felten, Raj, and Seamans linked AI benchmark performance to the abilities each occupation requires, later applying the same method to LLMs and finding language-heavy, cognitive occupations to be the most exposed \citep{felten_method_2018,felten_effect_2019,felten_occupational_2021,felten_how_2023}. Tomei, Teeselink, and Klein extend the approach beyond language models, assessing which tasks reinforcement learning could automate \citep{tomei2026jobs}. They find high exposure among monitoring and control occupations, such as gas plant operators and railroad conductors, that LLM-based measures overlook — a reminder that capability exposure depends on which technology is being measured. What these studies share is a focus on technical feasibility rather than on whether AI is used in any actual workplace. We group them under \textit{capability exposure}: the extent to which a task could be affected given what AI can technically do.

Capability alone does not put AI in a worker's hands. Technology can be adopted broadly only once it is built into software or an interface that workers can reach. Eloundou et al. accordingly measure not what an LLM alone could do, but how much an LLM paired with plausible software scaffolding could speed up a task, capturing potential access rather than raw capability \citep{eloundou2024gpts}. Chopra et al. map over 13,000 production-ready AI tools onto occupational skills, providing a measure of where AI capabilities are available through production-ready tools across the workforce \citep{chopra2025icebergindexmeasuringskillscentered}. Other studies infer availability from the supply and deployment of AI applications, using data on AI products developed by venture-funded startups \citep{Fenoaltea_2026}, AI deployment events \citep{demirev2026aiproduct}, or administrative data on AI applications \citep{desouza2025ai}. Together, these findings show that what AI reaches in practice is both smaller and differently distributed than what capability measures imply. We group these under \textit{availability exposure}: the extent to which a task is served by AI-enabled tools that already exist.

The availability of a tool does not ensure its use; a worker must judge it useful and worth adopting. Studies of usage logs reveal this gap directly. Massenkoff and McCrory find that actual LLM use covers far less of an occupation's activities than capability-based estimates predict, even in occupations those estimates rank as highly exposed \citep{anthropic2026aeiv5}. Where AI is used, moreover, that use clusters in a narrow range of activities: studies of Microsoft Copilot find information work to be the dominant case \citep{tomlinson_working_2025}, and analysis of Claude conversations shows a similar concentration in software development and writing \citep{handa2025economic}. We group these under \textit{observed exposure}: the extent to which AI is actually used for a task in practice.

Observed exposure thus comes closest of the three to real adoption, yet it still does not show whether AI has become part of how work is done. A usage record establishes that AI was used, not that a worker has come to rely on it. Two features of conversational logs make this ambiguity hard to resolve. First, sessions do not align neatly with tasks: one task may span several sessions, while one session may combine several tasks \citep{yang2026ai}. Second, a prompt requires little commitment, so the same log entry could reflect casual experimentation or genuine reliance \citep{handa2025economic}. Measuring how far AI has entered work therefore requires evidence of a different kind: evidence that a worker has chosen to delegate a task by building AI into a workflow. This is what we called \textit{delegated exposure} above, and develop its concept in the next section.

\subsection{Measuring delegated exposure through agentic adoption}
Practitioner-configured agents provide clearer evidence of intended delegation. Because current agentic systems lack independent goals, their behavior is determined by what the user specifies \citep{xing2026critiqueagentmodel}: a goal, a procedure for reaching it, and the tools or information needed for repeated execution \citep{11540994, gupta2026agenticaioccupationaldisplacement}. A configuration therefore records not merely that a practitioner used AI, but that they built it into how a task gets done. Delegation is therefore directly observable in agentic settings, whereas in conversational use it must be inferred from exchanges that rarely reveal whether a task was handed off or merely assisted.


Agentic systems also reach further into work than prompt-level LLM use. Rather than producing a single output, an agent shifts the practitioner from performing each step to supervising a whole sequence of work \citep{yang2026ai}. Despite this change, prior work has mainly focused on the capabilities of agents \citep{shao2026futureworkaiagents,gupta2026agenticaioccupationaldisplacement} or their availability \citep{sarkar2026enterprise}, rather than on delegated exposure. Studies that observe real usage logs, meanwhile, do not link them to the O*NET task structure \citep{johnston2026shiftagenticaievidence}, which makes their findings hard to compare with prior exposure measures. We therefore measure delegated exposure with the Agentic Adoption Index (AAI), built from practitioner-written descriptions of configured agents, linked to O*NET task statements, and aggregated to the occupation level. Because a configuration reflects a commitment already made rather than a possibility, delegated exposure should track where AI is most likely to change how work is done.

\section{Data}
\label{sec:data}

\subsection{Practitioner-configured agent skills}
\label{sec:data_agent}

Measuring delegated exposure calls for a different kind of data than the prompt-level conversation logs used in prior observed-exposure work. It requires evidence of how users configure AI agents to perform work, and such evidence should satisfy two conditions. First, users themselves should author it, so that it reflects their own configuration choices directly. Second, it should specify what an agent is intended to do in task-like terms, preferably at a level of abstraction comparable to occupational task descriptions as in O*NET.

The \texttt{skill} files published to public GitHub repositories\footnote{We draw on GitSkills, the first and only dataset to record the population of agent \texttt{skill} files in public GitHub repositories, where practitioners share them much as they share any open-source code---to distribute reusable automation recipes and to demonstrate what agentic workflows can accomplish. Rather than sampling, it captures the entire recorded population: $3{,}797{,}117$ \texttt{SKILL.md} files collected from $282{,}200$ public repositories, owned by $195{,}841$ accounts, in July 2026. The data is available at \url{https://doi.org/10.5281/zenodo.21875637}. These contributions come predominantly from technically sophisticated early adopters, yet span nearly the full range of occupations, from highly specialized tasks to routine ones. This breadth of openly shared, contributor-driven \texttt{skill}s is what makes the collection a useful window onto which occupational tasks practitioners are actually automating.} satisfy both conditions. Practitioners themselves author and publish these self-contained \texttt{SKILL.md} files. A \texttt{skill} is a modular resource that encapsulates a specific function or workflow, providing the agent with detailed instructions for performing specialized tasks.\footnote{Throughout this paper, we use typewriter font to denote this technical, platform-specific sense of the term, distinguishing it from the broader notion of ``skill'' used in the human capital literature.} Each \texttt{SKILL.md} file consists of a short description stating what the skill does and when it should be invoked, together with a body containing the full execution instructions. Because the description states what is done rather than how, it sits at the same level of abstraction as an O*NET task statement, which makes it a natural unit for semantic comparison with occupational tasks.

We collect accessible SKILL.md files from GitSkills, a dataset of skill files from public GitHub repositories [Destefanis et al., 2026], obtaining 3,797,117 SKILL.md files. Each SKILL.md file consists of YAML front matter specifying a natural-language \texttt{description}, followed by a body of execution instructions. We apply several filters to retain \texttt{skill}s usable for semantic matching. First, we remove \texttt{skill}s whose file content could not be retrieved, leaving $1{,}880{,}327$ \texttt{skill}s. We also remove \texttt{skill}s that contain a description but no execution instructions, as well as \texttt{skill}s whose description is under 100 characters, since such descriptions are too short to convey the automation context needed for a meaningful semantic match. This leaves $1{,}348{,}306$ \texttt{skill}s. Second, we restrict the sample to English-language descriptions using automated language identification \citep{lui-baldwin-2012-langid}. This leaves $1{,}190{,}508$ \texttt{skill}s. Third, we filter the remaining \texttt{skill}s for work-relatedness using GPT-4o-mini. We prompt the model to classify each description as work-related (a professional skill, workflow, tool usage, or occupational task) or not (unrelated to any work context, or describing only the setup of an agent without a concrete occupational task), retaining only the former (See SI Section~1 for the full prompt and settings). After removing duplicate descriptions, this retains $887{,}568$ \texttt{skill}s as our analysis sample.

\subsection{Occupational tasks and characteristics}
\label{sec:data_tasks}

To connect agent \texttt{skill}s to the occupations whose work they could automate, we need data on the tasks that make up each occupation's work. We draw on O*NET, a database developed and maintained by the US Department of Labor, using O*NET 30.2, released in 2025. It provides standardized, occupation-specific descriptions of the knowledge, skills, and tasks required across occupations, and remains the most widely used source for occupation-level task analysis. Among these, we focus on the task descriptors because agent \texttt{skill}s are themselves specifications of discrete tasks to be executed, making tasks the level at which the two can be directly compared. 

For each task, O*NET's Task Ratings report an importance score on a five-point scale, averaged across surveyed respondents. When aggregating task-level exposure scores to the occupation level, we use this mean importance as the weight for each task, so that tasks more central to an occupation contribute more to its occupation-level score. After excluding tasks without a valid importance rating, we retain 17,951 of the 18,797 O*NET task statements. This exclusion also removes 26 occupations whose tasks lack ratings entirely. When we compute AAI scores at the task level and aggregate them to the occupation level, we keep only the base O*NET-SOC codes (those ending in \texttt{.00}), dropping their finer subdivisions (e.g., \texttt{11-1011.03}), so that the occupation units align with the existing exposure measures and the wage and education data used in our analysis. This yields 748 occupations for analysis.

Finally, to examine how agentic adoption varies with occupational earnings and education, we draw on data on occupational wages, employment, and education from the Bureau of Labor Statistics Occupational Employment and Wage Statistics. For each occupation, we use the 2025 annual median wage, national employment level, and typical entry-level education requirement defined as the education level most commonly required to enter the occupation as designated by occupational experts. This information is available for 681 of the 748.

\subsection{Existing exposure measures}
\label{sec:data_exposure}

To understand how delegated exposure differs from previous measures, we compare it against existing measures from capability, availability, and observed exposure. We begin with a pre-AI baseline, the probability of computerisation \citep{frey2017future}, which estimates occupations' automation risk before large language models emerged and lets us test whether agentic diffusion follows the same routine-task patterns that earlier automation frameworks anticipated. We then turn to recent AI-exposure research and select measures according to three criteria. Each measure should (1) target current AI systems such as LLMs and agents, (2) adopt a distinct methodological approach spanning the capability, availability, and observed layers introduced in Section~\ref{sec:exposure}, and (3) report scores comparable at the occupation level. Guided by these criteria, we select three complementary measures, one per layer. The RL Feasibility Index of \citep{tomei2026jobs} captures capability exposure by scoring occupational tasks on their suitability for reinforcement-learning-based automation with Gemini 2.5 Flash. The task-level exposure ratings of \citep{eloundou2024gpts} capture availability exposure by assessing whether an LLM combined with complementary software could reduce task completion time by at least 50\% ($\gamma$). The usage-based measure of \citep{anthropic2026aeiv5} captures observed exposure based on actual Claude usage. All four measures, including the pre-AI baseline, are obtained directly from data released by their respective authors.

\section{Methods}
\label{sec:methods}

\subsection{Measuring task--skill similarity}
\label{sec:embedding}

\begin{figure*}[htbp]
  \centering
      \includegraphics[width=0.9\textwidth]{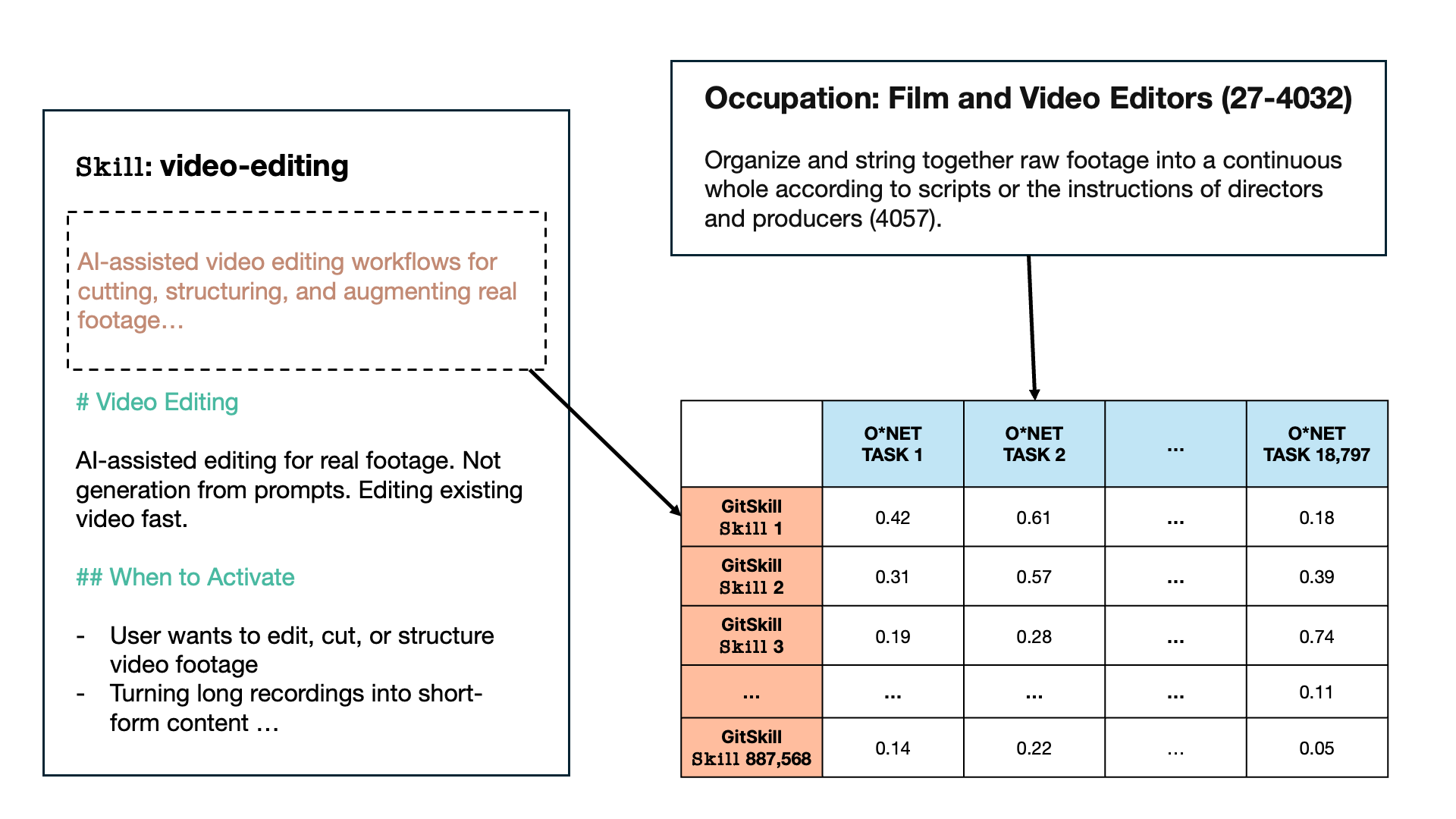}
  \caption{\label{fig:skill_structure}%
    \textbf{Example structure of a \texttt{skill.md} file and its mapping to O*NET tasks.}
    Each \texttt{skill} consists of a name, a short description, and a body containing full execution instructions. The description is embedded and compared against all O*NET task statements via cosine similarity, yielding a task-\texttt{skill} similarity matrix from which the occupation-level AAI is aggregated.}
\end{figure*}

Measuring how closely agent \texttt{skill}s correspond to occupational tasks requires a way to quantify semantic similarity between two large, unstructured collections of short text: the \texttt{skill} descriptions and the O*NET task statements. Embedding text into a shared vector space and measuring cosine similarity between texts is a common approach for this kind of matching in social science \citep{matsui2024word}. Following this approach, we embed every \texttt{skill} description and O*NET task statement into a common vector space using \texttt{all-MiniLM-L6-v2}, a Sentence-BERT model optimized for semantic similarity tasks that maps text to 384-dimensional dense vectors \citep{reimers2019sentence}. Because a \texttt{skill} often supports tasks spanning several occupations, we compute each \texttt{skill}'s semantic similarity against every O*NET task statement, allowing a single \texttt{skill} to match tasks across multiple occupations (see Figure~\ref{fig:skill_structure} for a schematic summary). We compute the cosine similarity between each task $t$ and \texttt{skill} $s$ as

\begin{equation}
  \label{eq:cosine}
  \text{sim}(t,s) = \frac{\bm{V}_{t} \cdot \bm{V}_{s}}{\|\bm{V}_{t}\|\,\|\bm{V}_{s}\|},
\end{equation}
where $\bm{V}_{t}$ and $\bm{V}_{s}$ are unit-normalized embedding vectors (See Tables S1 and S2 in SI Section 2 for the highest- and lowest-similarity tasks).

\subsection{Constructing the AAI}
\label{sec:AAI}
Having measured task--\texttt{skill} similarity, we aggregate these pairwise scores into a single index for each occupation. For each task $t$, we compute a task-level exposure score by averaging its cosine similarity across all \texttt{skill} descriptions in the corpus. We then aggregate to the occupation level using importance-weighted averaging over all tasks associated with occupation $j$:

\begin{equation}
  \label{eq:occadapt}
  \text{Agentic Adoption Index}_j = \sum_{t \in T_j} w_{jt} \cdot \frac{1}{|S|}\sum_{s \in S} \text{sim}(t, s),
\end{equation}


where $T_j$ is the set of O*NET task statements associated with occupation $j$, $S$ is the set of all retained \texttt{skill} descriptions with $|S| = 887{,}568$, and $w_{jt}$ is the importance weight of task $t$ within occupation $j$, normalized to sum to one so that occupations with more tasks are not scored higher by construction. A higher AAI indicates that an occupation's tasks are, on average, well covered by existing agentic \texttt{skill}s that early-adopting practitioners have already built and published. 

As a robustness check, we replicate the index using a different sentence embedding model, \texttt{all-mpnet-base-v2}, in place of \texttt{all-MiniLM-L6-v2}. The two occupation-level indices exhibit a Pearson correlation of 0.91, suggesting that the results are not sensitive to the choice of sentence embedding model.

\subsection{Validating the AAI}

We explore score highest and lowest on the AAI, which shows where agentic \texttt{skill}s align most and least closely with an occupation's tasks. Tables~\ref{tab:occ_top20} and \ref{tab:occ_bottom20} report the top and bottom 20 occupations by AAI. High-AAI occupations are information-intensive roles built around analysis, documentation, and coordination, such as \textit{management analysts}, \textit{technical writers}, and \textit{computer programmers}. Low-AAI occupations require manual dexterity or direct physical intervention, such as \textit{oral and maxillofacial surgeons}, \textit{roofers}, and \textit{tire builders}. To characterize this pattern, we relate the AAI to a cognitive ability share constructed from O*NET ability importance scores. The AAI is positively correlated with this share (See Figure S1 in SI Section 3), confirming that high-AAI occupations rely on cognitive rather than physical or psychomotor abilities.

\begin{table}[htbp]
\centering
\caption{Top 20 occupations by the Agentic Adoption Index}
\label{tab:occ_top20}
\footnotesize
\begin{tabular}{p{9cm}c}
\toprule
Occupation & AAI \\
\midrule
Computer Programmers & 0.191 \\
Software Quality Assurance Analysts and Testers & 0.187 \\
Management Analysts & 0.185 \\
Software Developers & 0.175 \\
Computer User Support Specialists & 0.175 \\
Technical Writers & 0.172 \\
Industrial Production Managers & 0.171 \\
Computer and Information Systems Managers & 0.171 \\
Database Administrators & 0.170 \\
Web Developers & 0.169 \\
Natural Sciences Managers & 0.168 \\
Database Architects & 0.168 \\
Production, Planning, and Expediting Clerks & 0.167 \\
Computer Systems Analysts & 0.166 \\
Architectural and Engineering Managers & 0.166 \\
Electronics Engineers, Except Computer & 0.164 \\
Computer Hardware Engineers & 0.162 \\
Statistical Assistants & 0.161 \\
Industrial Engineering Technologists and Technicians & 0.161 \\
Electrical and Electronic Engineering Technologists and Technicians & 0.160 \\
\bottomrule
\end{tabular}
\end{table}

\begin{table}[htbp]
\centering
\caption{Bottom 20 occupations by the Agentic Adoption Index}
\label{tab:occ_bottom20}
\footnotesize
\begin{tabular}{p{9cm}c}
\toprule
Occupation & AAI \\
\midrule
Oral and Maxillofacial Surgeons & 0.041 \\
Podiatrists & 0.045 \\
Tapers & 0.047 \\
Automotive Glass Installers and Repairers & 0.048 \\
Prosthodontists & 0.051 \\
Ophthalmic Medical Technicians & 0.055 \\
Dental Hygienists & 0.055 \\
Dermatologists & 0.057 \\
Roofers & 0.058 \\
Embalmers & 0.058 \\
Tire Builders & 0.060 \\
Shampooers & 0.061 \\
Dentists, General & 0.063 \\
Funeral Attendants & 0.063 \\
Slaughterers and Meat Packers & 0.063 \\
Carpet Installers & 0.063 \\
Animal Breeders & 0.064 \\
Bartenders & 0.064 \\
Cement Masons and Concrete Finishers & 0.064 \\
Bicycle Repairers & 0.065 \\
\bottomrule
\end{tabular}
\end{table}

\subsection{Analytic strategy}
\label{sec:analytic_strategy}
We examine the properties of AAI through three complementary analyses, each addressing a different question about what AAI captures. First, we ask whether the occupations that earlier, pre-AI frameworks rated as high in automation risk are the ones that now show high agentic adoption. We address this by comparing AAI with the pre-AI computerisation risk of \citep{frey2017future}. Second, we ask how agentic adoption relates to existing AI-exposure measures. These measures mostly capture general LLM use in services such as ChatGPT and Claude, but not agentic use. We address this by comparing AAI with three exposure measures spanning the capability, availability, and observed layers introduced in Section~\ref{sec:exposure}. Third, we ask how occupations at different wage and education levels adopt agentic AI, using employment-weighted least squares regressions. We explain each analytic approach in turn. 

\subsubsection{The AAI and pre-AI automation risk}
\label{sec:method_plane}
We examine whether the occupations that earlier, pre-AI frameworks rated as high in automation risk are the ones now adopting agentic AI most heavily. To answer this, we compare AAI with the probability of computerisation of \citep{frey2017future}, a widely cited occupation-level measure of automation risk. Because that measure classifies occupations as high- or low-risk, we split occupations at the median of AAI into high- and low-AAI groups. We then measure the overlap between the high-risk and high-AAI groups using the Jaccard similarity coefficient. A low Jaccard value would indicate that the occupations most exposed under pre-AI automation frameworks are largely not the ones adopting agentic AI most heavily today.


\subsubsection{The AAI and existing exposure measures}
\label{sec:method_validation}

Having introduced the AAI as a configuration-level extension of observed exposure, we compare it empirically with the capability, availability, and observed layers described in Section~\ref{sec:exposure}. We use the three exposure measures introduced in Section~\ref{sec:data_exposure}, one for each layer, to see which AAI tracks most closely. Table~\ref{tab:descriptives} reports summary statistics for these three measures and the AAI. The AAI has a mean and median of about 0.11 and a narrower range (0.04 to 0.19) than the others. The availability exposure $\gamma$ of \citep{eloundou2024gpts} spans the full 0 to 1 range with a median of 0.52, the Claude usage measure of \citep{anthropic2026aeiv5} concentrates near zero (median 0.00) with a long right tail up to 0.75, and the RL Feasibility Index of \citep{tomei2026jobs} ranges from 0 to about 69.42 with a median of 27.21.\footnote{The RL Feasibility Index is provided at the task level, and we aggregate it to the occupation level in the same way as the AAI.} These differences in scale motivate the rank-based comparisons used in the analyses that follow.

\begin{table}[htbp]
\centering
\caption{\label{tab:descriptives}%
  Descriptive statistics for the Agentic Adoption Index and existing
  exposure measures.}
\footnotesize
\begin{tabular}{lcccc}
\toprule
  & \shortstack{Agentic\\Adoption Index\\\mbox{}}&
   \shortstack{Tomei and Teeselink (2026)\\(Gemini)\\\textit{RL Feasibility Index}} &
   \shortstack{Eloundou et al. (2024)\\(GPT-4)\\\textit{Availability exposure} ($\gamma$)} &
   \shortstack{Massenkoff and McCrory (2026)\\(Claude)\\\textit{Claude usage measure}} \\
\midrule
$N^{\dagger}$ & 748   & 894    & 798   & 756   \\
Mean     & 0.11  & 26.78  & 0.51  & 0.08  \\
Min      & 0.04  & 0.00   & 0.00  & 0.00  \\
25th pct & 0.09  & 14.63  & 0.17  & 0.00  \\
Median   & 0.11  & 27.21  & 0.52  & 0.00  \\
75th pct & 0.12  & 36.86  & 0.85  & 0.10  \\
Max      & 0.19  & 69.42  & 1.00  & 0.75  \\
\bottomrule
\multicolumn{5}{p{\textwidth}}{\footnotesize $^{\dagger}$Occupation counts vary slightly across measures because each source relies on a different O*NET release and level of measurement (e.g., tasks, detailed work activities, or skill-level descriptions), which map to occupations with slightly different coverage.} \\
\end{tabular}
\end{table}

\subsubsection{The AAI, wages, and education}
\label{sec:method_regression}
We examine how occupations at different wage and education levels adopt agentic AI through a series of employment-weighted least squares regressions with the occupation-level AAI as the dependent variable. We weight each occupation by its 2025 employment, so that the results describe patterns across the workforce rather than across job titles. Model~(1) asks whether the AAI rises with wages and how it varies across education levels, reaching its highest values in the best-paid and most-educated occupations. We regress the AAI of occupation \textit{j} on the logarithm of the 2025 median annual wage $w_j$, together with education dummies for high school or below ($D^{HS}_j$) and master's or above ($D^{MA}_j$), relative to a bachelor's degree,
\begin{equation}
\label{eq:model1}
\text{Model (1): }\text{AAI}_j = \beta_0 + \beta_1 \log w_j + \beta_2 D^{HS}_j + \beta_3 D^{MA}_j + \varepsilon_j.
\end{equation}
Model~(2) turns to the availability of AI, asking whether occupations whose tasks current AI tools can already handle are also those for which practitioners have built the most agent \texttt{skill}s. We regress the AAI on the availability exposure $\gamma_j$ of \citep{eloundou2024gpts} alone,
\begin{equation}
\label{eq:model2}
\text{Model (2): }\text{AAI}_j = \beta_0 + \beta_1 \gamma_j + \varepsilon_j.
\end{equation}
Model~(3) brings the two together, examining whether the association between the AAI and $\gamma_j$ remains significant once wages and education are added as controls, and, conversely, whether wages and education remain significant once $\gamma_j$ is held constant,
\begin{equation}
\label{eq:model3}
\text{Model (3): }\text{AAI}_j = \beta_0 + \beta_1 \gamma_j + \beta_2 \log w_j + \beta_3 D^{HS}_j + \beta_4 D^{MA}_j + \varepsilon_j.
\end{equation}

Finally, Model~(4) considers that the way AAI adoption rises with wages may itself differ across education levels. We interact the log wage with each education dummy, so that the wage slope is allowed to vary among occupations requiring high school or below, a bachelor's degree, and a master's or above,

\begin{equation}
\label{eq:model4}
\begin{split}
\text{Model (4): }\text{AAI}_j = {}& \beta_0 + \beta_1 \gamma_j + \beta_2 \log w_j + \beta_3 D^{HS}_j + \beta_4 D^{MA}_j \\
& + \beta_5 (\log w_j \times D^{HS}_j) + \beta_6 (\log w_j \times D^{MA}_j) + \varepsilon_j.
\end{split}
\end{equation}
With these four models, we compare the coefficients across them and interpret the differences.

\section{Results}
\label{sec:results}
\subsection{Agentic adoption does not follow pre-AI automation risk}
\label{sec:plane}
\begin{figure*}[htbp]
  \centering
  \includegraphics[width=0.7\textwidth]{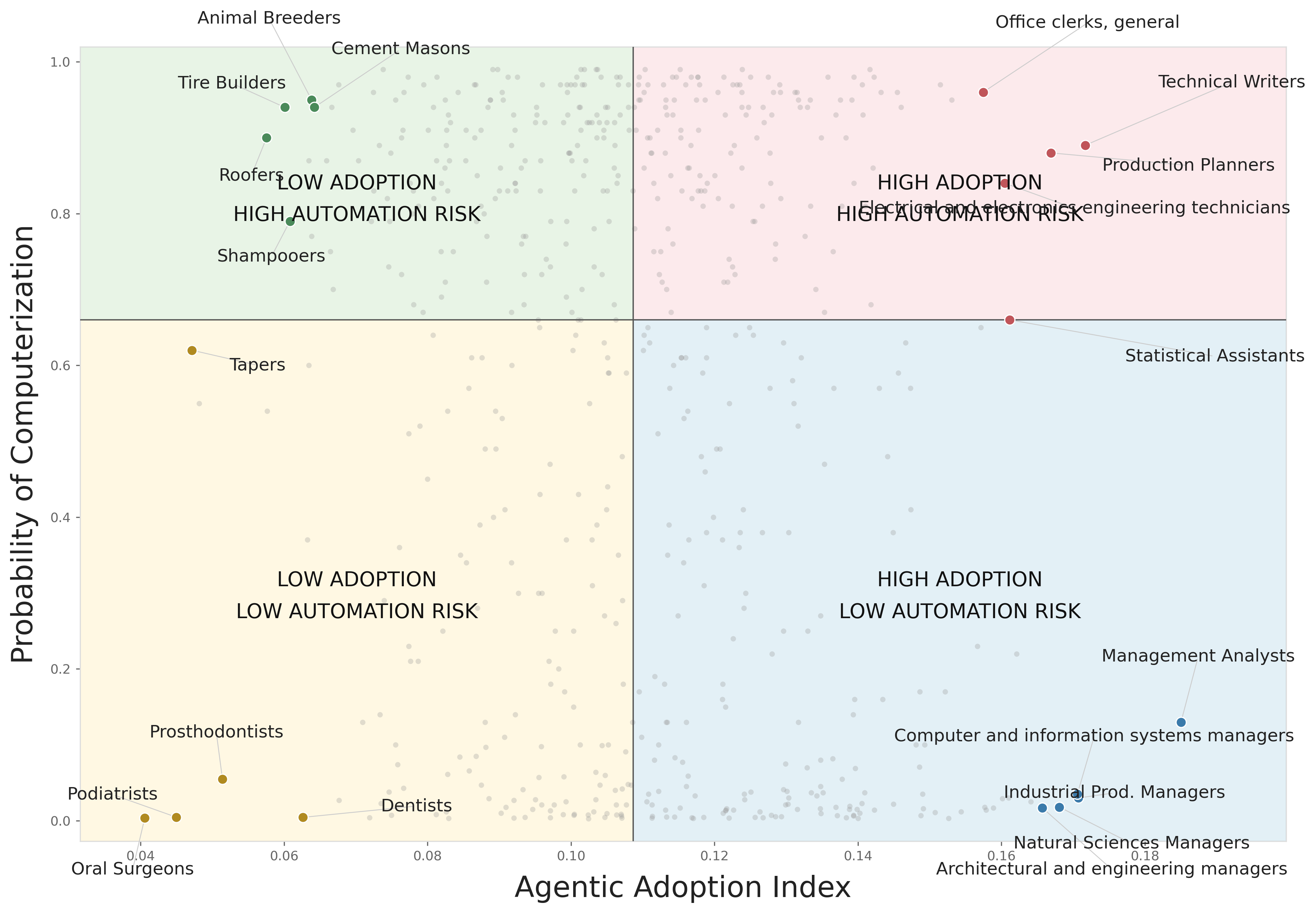}
  \caption{\textbf{AAI versus Computerization Risk across occupations.}
    The $x$-axis is AAI; the $y$-axis is the probability of computerization from \cite{frey2017future}. Occupations with similar computerization risk diverge widely in the AAI, so the two measures capture different sets of occupations.}
  \label{fig:quadrant}
\end{figure*}

We compare the AAI with the pre-AI automation risk of \citep{frey2017future} to assess whether the occupations now adopting agentic AI are those that earlier frameworks expected to be automated. Plotted against the probability of computerisation, the AAI shows little correspondence with it (Figure~\ref{fig:quadrant}). Occupations that earlier frameworks grouped together by automation risk span a wide range of AAI values. Among occupations with low pre-AI automation risk, some score low on AAI, such as \textit{oral surgeons}, \textit{prosthodontists}, and \textit{podiatrists}, while others score high, such as \textit{management analysts}, \textit{industrial production managers}, and \textit{natural sciences managers}. Occupations with high pre-AI automation risk divide in the same way, with manual roles such as \textit{tapers} and \textit{tire builders} scoring low and information-intensive ones such as \textit{technical writers} and \textit{statistical assistants} scoring high.

To quantify this divergence, we split occupations at the median of AAI into high and low groups, mirroring the high- and low-risk partition of the Frey--Osborne measure, and assess the agreement between the two classifications with the Jaccard similarity coefficient. The coefficient is only $0.33$, indicating that the occupations flagged as high-risk by the pre-AI measure are for the most part distinct from those now exhibiting the highest AAI. This weak overlap between pre-AI risk and AAI also holds at the sector and state levels (See Figures S2-S4 in SI Section 4). 

\subsection{The AAI aligns with capability more than with observed use}
\label{sec:validation}

\begin{figure*}[htbp]
  \centering
  \includegraphics[width=\textwidth, keepaspectratio]{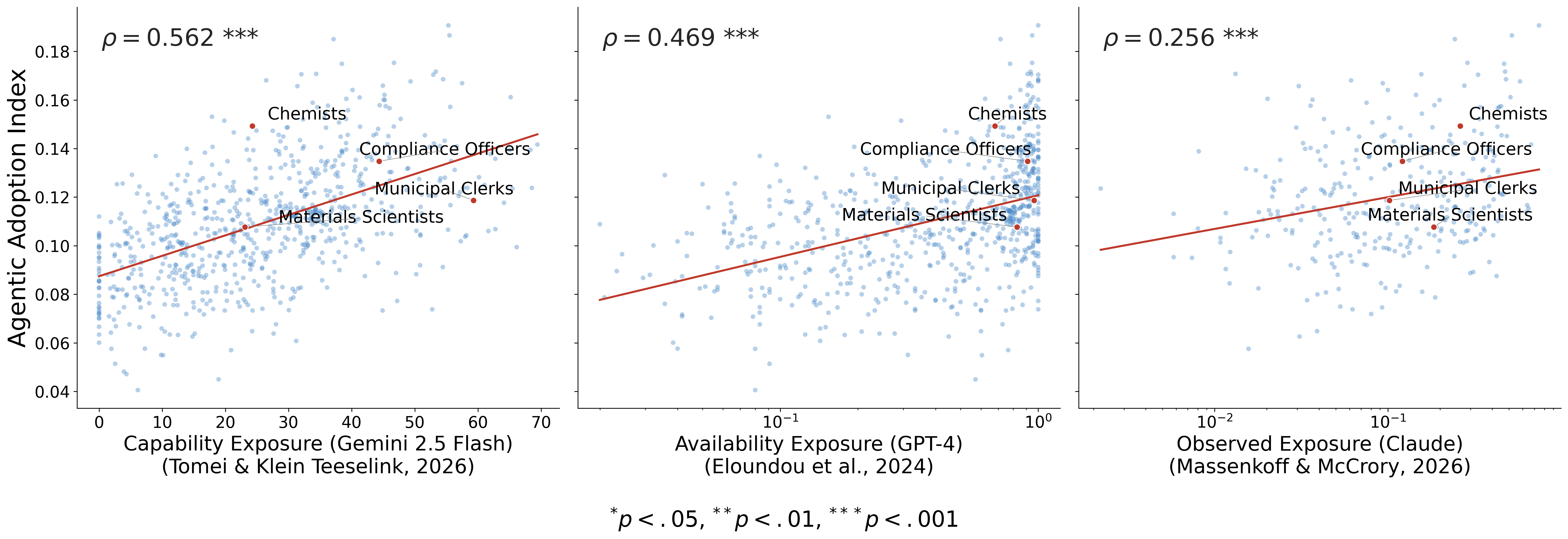}
  \caption{\textbf{Correlation between AAI and established exposure measures.}
    Each panel plots the AAI against one of three occupation-level exposure scores: the Gemini-based capability exposure (RL Feasibility Index) from \citep{tomei2026jobs} (left), ChatGPT-based availability exposure $\gamma$ from \citep{eloundou2024gpts} (center), and Claude-based observed exposure from \citep{anthropic2026aeiv5} (right). The regression line is fitted in linear space for the left panel and in log-$x$ space for the center and right panels.}
  \label{fig:validation}
\end{figure*}

Because the AAI introduces a configuration-level view of exposure, we situate it relative to established measures, each of which captures a distinct layer of how much AI can affect an occupation. We compare the AAI against the three measures of Section~\ref{sec:data_exposure}, one per layer, to identify the layer to which agentic adoption corresponds most closely. All three Spearman correlations are positive and significant (Figure~\ref{fig:validation}), indicating that our practitioner-derived index is broadly consistent with prior work. The correlations differ in strength, however. The AAI correlates most strongly with the capability measure of \citep{tomei2026jobs} ($\rho = 0.562^{***}$), followed by the availability measure of \citep{eloundou2024gpts} ($\rho = 0.469^{***}$), whereas its correlation with the observed usage measure of \citep{anthropic2026aeiv5} ($\rho = 0.256^{***}$) is considerably weaker. The AAI thus aligns more closely with AI capability and availability than with where it is already deployed in practice. This ordering across the three measures suggests that, early in the diffusion of agentic AI, the AAI captures the potential for agents to take over an occupation's tasks rather than the extent to which that occupation currently uses AI.

\subsection{Adoption rises with wages, except among the most-educated}
\label{sec:regression_results}

To see how agentic adoption is positioned across the labor market, we examine the regression results from Section~\ref{sec:method_regression} in Table~\ref{tab:regression}. Model~(1) shows that the AAI increases with wages: a higher log wage is associated with a higher AAI ($\beta = 0.036^{***}$), so better-paid occupations tend to adopt agentic AI more. When it comes to education, occupations requiring high school or below do not differ significantly from bachelor's-degree occupations ($\beta = 0.002$), whereas those requiring a master's or above have significantly lower AAI ($\beta = -0.031^{***}$). The AAI is thus no higher for occupations requiring high school or below than for bachelor's-degree occupations, and significantly lower only for those requiring a master's or above.


Model~(2) relates the AAI to the availability of AI to see whether users adopt AI agents when the service becomes commercialized. On its own, the availability exposure $\gamma$ enters positively and significantly ($\beta = 0.048^{***}$) and alone explains $44\%$ of the variance in AAI (Adj. $R^{2} = 0.442$), more than wages and education together in Model~(1) (Adj. $R^{2} = 0.346$). Agentic adoption is thus more strongly associated with whether current AI tools can handle an occupation's tasks than with its wage or education levels.


Model~(3) includes wages, education, and $\gamma$ together. The availability exposure $\gamma$ remains significant ($\beta = 0.038^{***}$), though smaller than on its own in Model~(2) ($0.048^{***}$), because occupations with higher AI availability also tend to have higher average wages and education levels. The wage coefficient also retains significance ($\beta = 0.022^{***}$), down from $0.036^{***}$ in Model~(1). The positive wage relationship therefore holds but weakens once $\gamma$ is controlled for, indicating that the initial wage effect in Model~(1) was partially driven by underlying AI availability.


For occupations requiring a master's degree or above, the gap remains significant and changes little, from $-0.031^{***}$ to $-0.025^{***}$, so availability alone does not account for their lower AAI. For occupations requiring high school or below, the coefficient is not significant in Model~(1) ($0.002$) but turns significantly positive once availability is controlled ($0.008^{***}$ in Model~(3)). Holding availability fixed, then, the AAI appears higher the lower an occupation's required education. This pattern, however, assumes the AAI responds to wages the same way at every education level; Model~(4) tests this assumption by letting the wage slope differ by education, and finds that it may not hold.


Model~(4) asks whether the AAI rises with wages differently across education levels. Among bachelor's-degree occupations, the AAI continues to rise with wages ($\beta = 0.036^{***}$). For occupations requiring high school or below, the interaction with the log wage is small but significantly negative ($\beta = -0.013^{***}$), so their AAI still rises with wages, though somewhat less steeply than at the bachelor's level. For those requiring a master's or above, by contrast, the interaction is far larger and negative ($\beta = -0.055^{***}$). It more than cancels out the positive wage effect, so their AAI no longer rises with wages and instead declines. The wage relationship thus holds up to the bachelor's level but reverses among the most-educated occupations. Adding the interaction terms raises Adj. $R^{2}$ modestly, from $0.526$ to $0.554$.

\begin{figure*}[htbp]
  \centering
\includegraphics[width=\textwidth]{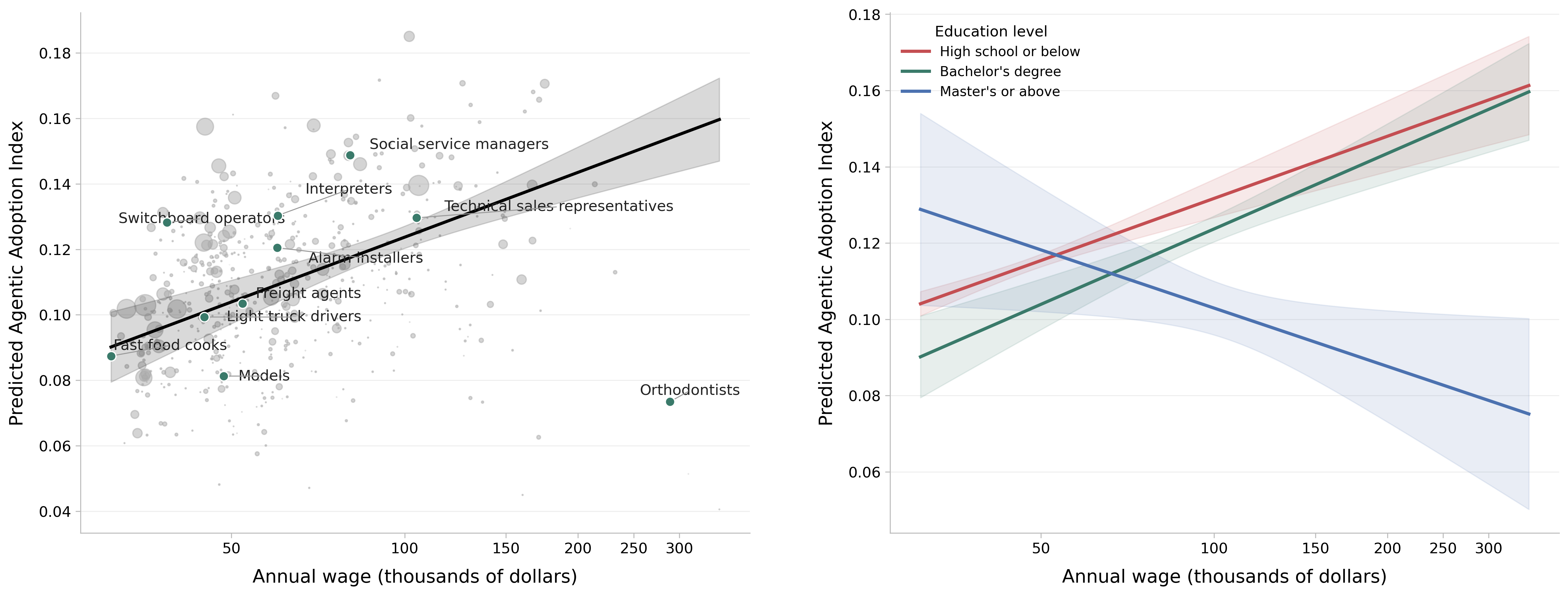}
  \caption{\textbf{The AAI's relationships with wage and education.}
    Both panels are based on the full WLS regression in Model (4) of Table~\ref{tab:regression}, which includes an interaction between $\log$ wage and education.
    (Left) Predicted AAI across annual wage, holding education at the Bachelor's degree reference group and $\gamma$ at its employment-weighted mean. Each point is an occupation, with dot size proportional to employment, and the solid line is the model-predicted AAI with the shaded band denoting its 95\% confidence interval.
    (Right) Predicted AAI across annual wage for each education level, holding $\gamma$ at its employment-weighted mean. Solid lines are the model-predicted AAI for each group, shaded bands denoting their 95\% confidence intervals. The diverging slopes reflect the interaction between $\log$ wage and education: predicted AAI rises with wage among the high school and Bachelor's groups but declines among the Master's-or-above group.}
  \label{fig:wage}
\end{figure*}

Figure~\ref{fig:wage} illustrates this pattern, with both panels based on Model~(4). The left panel, with education fixed at the bachelor's level, shows that the predicted AAI increases with the annual wage. The right panel plots the predicted AAI separately for each education group. For occupations requiring a bachelor's degree or less, the predicted AAI increases with the annual wage, with little substantive difference between the two groups. For those requiring a master's or above, by contrast, it slopes downward, falling below the other two groups at high wages.


\begin{table*}[htbp]
\centering
\caption{\label{tab:regression}%
  WLS regressions of the Agentic Adoption Index.
  Standard errors in parentheses.}
{\small
\setlength{\tabcolsep}{10pt}
\begin{tabular}{lcccc}
\toprule
  & (1) & (2) & (3) & (4) \\
\midrule
Intercept
  & $-0.280^{***}$ & $0.090^{***}$ & $-0.151^{***}$ & $-0.305^{***}$ \\
  & $(0.031)$ & $(0.001)$ & $(0.028)$ & $(0.047)$ \\[6pt]
Availability Exposure ($\gamma$)
  & & $0.048^{***}$ & $0.038^{***}$ & $0.038^{***}$ \\
  & & $(0.002)$ & $(0.002)$ & $(0.002)$ \\[6pt]
\midrule
$\log$ wage
  & $0.036^{***}$ & & $0.022^{***}$ & $0.036^{***}$ \\
  & $(0.003)$ & & $(0.002)$ & $(0.004)$ \\[4pt]
High school or below
  & $0.002$ & & $0.008^{***}$ & $0.160^{***}$ \\
\quad(ref.\ Bachelor's)
  & $(0.003)$ & & $(0.002)$ & $(0.057)$ \\[4pt]
Master's or above
  & $-0.031^{***}$ & & $-0.025^{***}$ & $0.612^{***}$ \\
\quad(ref.\ Bachelor's)
  & $(0.004)$ & & $(0.003)$ & $(0.096)$ \\[4pt]
$\log$ wage $\times$ High school or below
  & & & & $-0.013^{***}$ \\
  & & & & $(0.005)$ \\[4pt]
$\log$ wage $\times$ Master's or above
  & & & & $-0.055^{***}$ \\
  & & & & $(0.008)$ \\[6pt]
\midrule
$R^{2}$
  & $0.349$ & $0.443$ & $0.529$ & $0.558$ \\
Adj.\ $R^{2}$
  & $0.346$ & $0.442$ & $0.526$ & $0.554$ \\
$N$
  & $681$ & $681$ & $681$ & $681$ \\
\bottomrule
\multicolumn{5}{l}{$^{*}p<.1$, $^{**}p<.05$, $^{***}p<.01$} \\
\end{tabular}}
\end{table*}


Alternative exposure measures show a similar pattern (see Table S3 in SI Section~5). The RL Feasibility Index of \citep{tomei2026jobs}, a capability-type measure, explains about $58\%$ of the variance in AAI once wages and education are added (Adj.\ $R^{2} = 0.584$), close to the availability exposure used in the main models. The Claude usage measure of \cite{anthropic2026aeiv5}, an observed-type measure, explains less even with education included (Adj.\ $R^{2} = 0.503$). The AAI therefore reflects what early adopters have already built as agentic workflows, rather than how widely AI is used across the broader workforce. Across every specification that includes them, the log wage remains positive and significant, and its interaction with a master's degree or above stays significantly negative ($-0.052^{***}$ to $-0.067^{***}$). Its interaction with high school or below is significantly negative only under the capability measure ($-0.012^{**}$) and not under the availability or observed measures ($-0.006$ and $0.007$, respectively).

Taken together, the four models point to one conclusion. People delegate their work to agents mostly when current AI tools can already handle it. Wages and education matter too, but mainly in how they interact: the wage relationship holds across most occupations yet breaks down among the most-educated. The AAI does rise with wages, but this is partially because better-paid occupations tend to be ones that current AI tools can already reach; once availability is accounted for, the wage relationship weakens. This wage relationship, however, varies with education: for occupations requiring a master's degree or above, the AAI no longer rises with wages. These occupations adopt agentic AI less than others of comparable availability, and this gap widens as their wages rise.

\subsection{Robustness to an alternative skill corpus}
\label{sec:robustness_manus}
The results above are built from a single corpus of agent skills. To check that they do not depend on this particular source, we reconstruct the AAI from an entirely separate corpus, the Manus Skills Marketplace, and repeat the analyses of Sections~\ref{sec:plane}--\ref{sec:regression_results}. Although the Manus corpus is considerably smaller (roughly 53,000 skills, compared with 888,000 in GitSkills) and was collected independently, the Manus-based index yields results closely consistent with those reported above (see Section~6 of the Supplementary Information for full results).

\section{Discussion and Conclusion}
\label{sec:discussion}

We extend research on AI exposure by adding the concept of delegated exposure, as LLM use moves beyond isolated conversations toward reusable agentic workflows. We operationalize it through the AAI, which matches skill descriptions from the public GitHub repositories to O*NET task statements. The resulting measure reveals a distinct occupational gradient. Delegated exposure concentrates in information-intensive work and remains minimal where work depends on manual dexterity or direct intervention in the physical environment. This distribution departs from pre-AI accounts such as computerization-risk estimates, which located automation risk in routine tasks and ranked occupations differently from what we observe. The AAI also aligns more closely with measures of technical capability and application availability than with prompt-level conversational use, suggesting that it tracks what AI can in principle perform rather than what usage records currently capture. Yet the availability measure does not explain the pattern fully: adoption increases with wages among occupations requiring a bachelor's degree or less, but declines among higher earners requiring advanced degrees. These patterns hold when the AAI is reconstructed from an independently collected corpus of agent skills from the Manus Skills Marketplace.

Our findings show that occupational AI exposure needs to be interpreted in relation to the stage and form of diffusion under examination. Measures of capability, availability, and observed exposure are not interchangeable readings of a single underlying quantity; each captures a distinct phase, from what a technology can do, to what has been built with it, to what people take up. Disagreements in this literature may therefore partly reflect measures of different layers being compared as though they were the same thing.

The AAI adds a further layer by identifying where practitioners have begun configuring AI systems to automate their own work, and its divergence from observed exposure is telling. A prompt records a single request, whereas an agent description specifies the role, procedure, and sequence of activities a practitioner intends to automate. Because that specification takes effort, practitioners undertake it only where the technology can dependably do the work, which is why delegated exposure resembles capability exposure more than observed exposure does. Conversational records, in turn, may understate how far practitioners have committed to automating work.

Our models show that technical capability accounts for most of the variation in agent adoption than availability or observed exposure. This complicates a well-established expectation that use lags capability: general purpose technologies typically deliver returns only after firms make the complementary investments in process redesign, training, and reorganization that let the technology be absorbed \citep{taylor2009organizational, brynjolfsson2013complementarity, machkour2026artificial}, and technologies are reshaped by the settings that take them up \citep{klein2002social, berker2005domestication}. Our results suggest those adjustments may play a smaller role than expected, at least in the early adoption of agentic systems.

Two features of agentic systems may explain this pattern. First, these systems are configured in natural language, so for the practitioners in our data the distance between recognizing that a task could be automated and building something that does it is short. The skill files we observe were written and published by individuals, without the procurement, integration, or specialized expertise that would ordinarily mediate a workplace technology \citep{bright2025widespread}. Second, generality means the technology adapts to the task rather than the task to the technology, removing much of the mutual adjustment that typically slows adoption. 

Yet this pattern is not uniform. The AAI exhibits a distinct non-linear pattern across wages and education: adoption increases alongside wages for occupations requiring a bachelor's degree or less, but declines among higher earners requiring a graduate degree. Technical availability partially explains this pattern, however, the overall trend holds: availability alone cannot account for the lower adoption observed among the most educated and highest-paid occupations. One explanation is that work at the top depends on tacit expertise, contextual judgment, and intensive interpersonal interaction, none of which is easily specified in advance \citep{autor2015there}. But our data cannot distinguish constraint from choice. These occupations may be unable to specify their work in advance, or they may be able to and decline, since professionals have both the autonomy and the incentive to control the pace at which their work is codified \citep{abbott1988system}. The AAI therefore captures not where automation has already occurred, but where early practitioners are actively attempting to automate their own work. 

Longitudinal evidence would help here, and not only here. If the shortfall at the top narrows as agentic systems improve, constraint is the better explanation; if it persists while capability rises, the case for choice strengthens. Also, pre-AI computerisation estimates were influential within a decade of publication and are already dated, which suggests that any single-period measure of exposure has a short shelf life. Tracking each layer over time would show not only how exposure grows but how the layers move in relation to one another.

Sustaining such tracking over time faces a structural obstacle. Conversational usage data sits with a small number of platform firms, and expert annotation of tasks is costly and slow to update. Publicly shared artifacts avoid both constraints. The skill files we analyze from GitHub were published by practitioners for other practitioners, which makes them observable without platform cooperation and renewable as the number of participants grows. Our data source suggests that the traces left by users configuring their own tools may provide a more accessible basis for tracking diffusion over time.

Several limitations qualify these conclusions, suggesting directions for future research. First, the agent corpus likely overrepresents technically sophisticated, highly engaged practitioners, so the AAI should be read as an indicator of early-adopter activity rather than of adoption across the workforce. Tracking it longitudinally as practices mature would show whether the present concentration persists or spreads to other areas of work. Second, the AAI captures whether practitioners have configured an agent for a task, not how the agent structures and carries out the work. An agent may decompose or sequence a task in ways that differ substantially from how a human performs it. Examining execution traces or the internal structure of agent runs, rather than skill descriptions alone, would clarify how agentic task performance diverges from human work. Third, the analysis relies on a static list of O*NET tasks. As \citep{acemoglu2019automation} note, the full effect of automation involves both a displacement effect, in which machines substitute for existing tasks, and a reinstatement effect, in which new tasks emerge. The AAI speaks only to the former and cannot represent the new work that agents create. Capturing reinstatement will require task inventories that update as new work emerges, drawn from sources such as job postings or agent operation records. Such an approach would reveal not only where work is being automated but also where new forms of human work are being created.


\subsection*{Data and code availability} All data and replication code are available at (anonymized for peer-review).


\clearpage
\bibliographystyle{plainnat}

\bibliography{references_ver2.0}

\end{document}